\documentclass[twocolumn, switch]{article} 

\usepackage{arxiv}

\usepackage{amsmath, amsthm, amssymb, amsfonts}
\usepackage{algpseudocode}
\usepackage{algorithm}

\usepackage[numbers,square]{natbib}
\usepackage[utf8]{inputenc}	
\usepackage[T1]{fontenc}	
\usepackage{xcolor}		
\usepackage[colorlinks = true,
            linkcolor = purple,
            urlcolor  = blue,
            citecolor = cyan,
            anchorcolor = black]{hyperref}	
\usepackage{booktabs} 		
\usepackage{nicefrac}		
\usepackage{microtype}		
\usepackage{lineno}		
\usepackage{float}			

\usepackage{lipsum}		

\usepackage{newfloat}
\DeclareFloatingEnvironment[name={Supplementary Figure}]{suppfigure}
\usepackage{sidecap}
\sidecaptionvpos{figure}{c}

\usepackage{titlesec}
\titlespacing\section{0pt}{12pt plus 3pt minus 3pt}{1pt plus 1pt minus 1pt}
\titlespacing\subsection{0pt}{10pt plus 3pt minus 3pt}{1pt plus 1pt minus 1pt}
\titlespacing\subsubsection{0pt}{8pt plus 3pt minus 3pt}{1pt plus 1pt minus 1pt}

\usepackage{tikz,xcolor,hyperref}

\definecolor{lime}{HTML}{A6CE39}
\DeclareRobustCommand{\orcidicon}{
	\begin{tikzpicture}
	\draw[lime, fill=lime] (0,0) 
	circle [radius=0.16] 
	node[white] {{\fontfamily{qag}\selectfont \tiny ID}};
	\draw[white, fill=white] (-0.0625,0.095) 
	circle [radius=0.007];
	\end{tikzpicture}
	\hspace{-2mm}
}
\foreach \x in {A, ..., Z}{\expandafter\xdef\csname orcid\x\endcsname{\noexpand\href{https://orcid.org/\csname orcidauthor\x\endcsname}
			{\noexpand\orcidicon}}
}

\title{MathNet: A Data-Centric Approach for Printed Mathematical Expression Recognition}

\usepackage{authblk}

\author[1, 2\thanks{\tt{scmx@zhaw.ch}}]{Felix M. Schmitt-Koopmann\orcidA{}}
\author[2]{Elaine M. Huang\orcidB{}}
\author[1]{Hans-Peter Hutter\orcidC{}}
\author[3,4]{\\Thilo Stadelmann\orcidD{}}
\author[1]{Alireza Darvishy\orcidE{}}

\affil[1]{Institute of Computer Science, ZHAW, 8401 Winterthur, Switzerland}
\affil[2]{People and Computing Laboratory, University of Zurich, 8050 Zurich, Switzerland}
\affil[3]{Centre for Artificial Intelligence, ZHAW, 8400 Winterthur, Switzerland}
\affil[4]{European Centre for Living Technology (ECLT), 30123 Venice, Italy}

\begin{document}

\twocolumn[ 
  \begin{@twocolumnfalse} 
  
\maketitle

\begin{abstract}
Printed mathematical expression recognition (MER) models are usually trained and tested using LaTeX-generated mathematical expressions (MEs) as input and the LaTeX source code as ground truth. As the same ME can be generated by various different LaTeX source codes, this leads to unwanted variations in the ground truth data that bias test performance results and hinder efficient learning. In addition, the use of only one font to generate the MEs heavily limits the generalization of the reported results to realistic scenarios. We propose a data-centric approach to overcome this problem, and present convincing experimental results: Our main contribution is an enhanced LaTeX normalization to map any LaTeX ME to a canonical form. Based on this process, we developed an improved version of the benchmark dataset \emph{im2latex-100k}, featuring $30$ fonts instead of one. Second, we introduce the real-world dataset \emph{realFormula}, with MEs extracted from papers. Third, we developed a MER model, \emph{MathNet}, based on a convolutional vision transformer, with superior results on all four test sets (\emph{im2latex-100k}, \emph{im2latexv2}, \emph{realFormula}, and \emph{InftyMDB-1}), outperforming the previous state of the art by up to $88.3$\%.
\end{abstract}
\vspace{0.35cm}

  \end{@twocolumnfalse} 
] 

\section{Introduction}
\label{sec:Introduction}
\begin{figure}
    \centering
    \includegraphics[width=\linewidth]{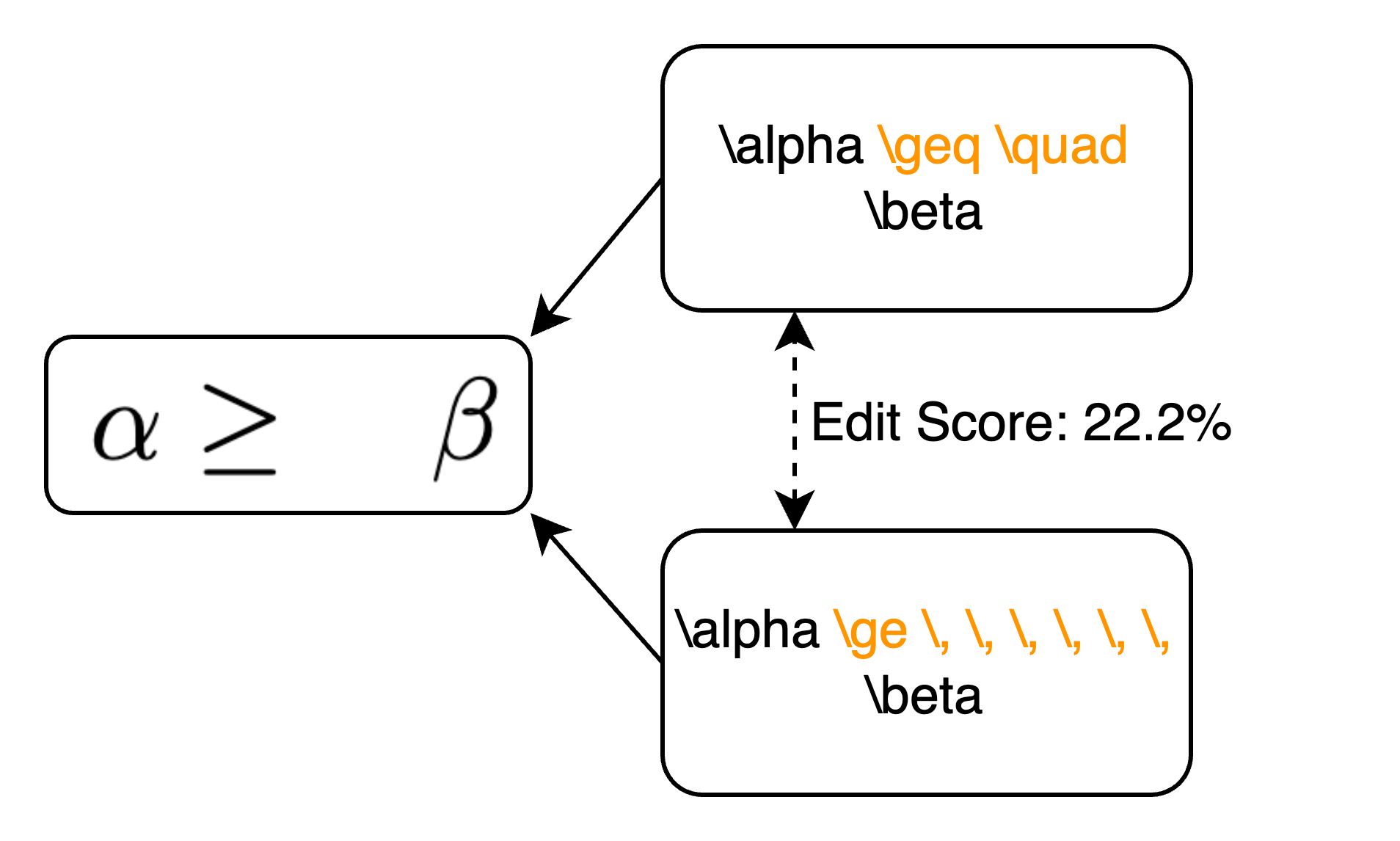}
    \caption{An example of an ME image that can be produced with more than one LaTeX code. While the two presented LaTeX codes are quite different ($22.2$\% Edit score), they create the same image.}
    \label{fig:LaTeXproblem}
\end{figure}

Recognizing mathematical expressions (MEs) in images and converting them into a machine-understandable format is known as mathematical expression recognition (MER). Creating a dependable MER would unlock possibilities for producing innovative tools, such as the ability to digitize, search, extract, and enhance the accessibility of mathematical equations in documents \cite{schmitt-koopmann_accessible_2022}.\\
However, despite recent progress in the field of MER, it remains a challenge for two main reasons. Firstly, MEs contain many symbols, i.e., multiple alphabets, numerals, operators, and parentheses. Secondly, structural information (for example, nested superscripts and subscripts) is crucial for correctly recognizing MEs \cite{belaid_syntactic_1984,chan_mathematical_2000}.\\
In addition, we have identified a third challenging aspect that needs to be addressed. The machine-understandable format used in many MER models can cause unwanted variation. For instance, LaTeX is a popular format used by many MER models \cite{aggarwal_survey_2022}. However, LaTeX allows authors to write the same ME with different LaTeX code as shown in Figure \ref{fig:LaTeXproblem}. Accordingly, many LaTeX commands are redundant or can be neglected without altering the canonical form or even without changing the visual appearance of an ME. For example, we observed that of the $500$ different tokens in the printed MER benchmark dataset im2latex-100k \cite{kanervisto_im2latex-100k_2016}, $174$ tokens or $34.8$\% of the vocabulary is redundant or does not influence the canonical form of the ME. This leads to detrimental variability in the training data and therefore to inefficient learning, excessive training data needs, and, finally, suboptimal recognition performance due to unresolved ambiguity in the model\cite{zhang_understanding_2021, arpit_closer_2017}. Finally, the use of a single font for MEs in \emph{im2latex-100k} heavily limits the generalization of performance results reported on this data set to realistic scenarios.\\
Reducing this variability is not only to reduce unwanted biases in test scores but is expected to have a high impact on learning quality of respective models and, hence, their performance. Recently, methods have emerged under the term data-centric AI that systematically engineer the data in order to improve overall system quality \cite{deeplearningai_chat_2021, luley_concept_2023}. The methodology is characterized by making the data a first-class citizen in the development process of any machine-learning-based system, thus shifting the focus away from merely manipulating the model architecture \cite{stadelmann_data_2022}. 
In this paper, we adopt a data-centric approach by proposing a systematic process to map an ME to a canonical LaTeX representation. Since we develop our methods to make formulae in PDFs accessible, we focus in this work on printed MER, but our approaches are also applicable to handwritten MEs and we expect similar benefits for complex handwritten MEs. We present the following major contributions: 1) A LaTeX normalization process that maps LaTeX MEs to a canonical form. 2) \emph{Im2latexv2}, an upgraded version of \emph{im2latex-100k} with multiple fonts and a canonical ground truth (GT). 3) \emph{realFormula}, a real-world test set for MER. 4) Our MER model \emph{MathNet} which outperforms the previous state of the art on all four test sets by up to $88.3$\%. \\
The remainder of this paper is structured as follows: we present related work in Section \ref{sec:RelatedWork}. We discuss the issues with using LaTeX for MER in Section \ref{sec:ProblematicLaTeX}. We introduce the datasets we have developed, the metrics used, and our printed-MER model in Section \ref{sec:ApplyingLaTeXNormalizationforPrintedMER}. We will then present the results of our experiments in Section \ref{sec:experiments} and discuss them in Section \ref{sec:discussion}. Finally, we offer concluding remarks in Section \ref{sec:conclusion}.

\section{Related Work}
\label{sec:RelatedWork}
MER has been a research task for over $50$ years \cite{anderson_syntax-directed_1967}, and it still remains open. Although the focus of the MER research field has shifted to the recognition of online and offline handwritten MEs in the last years, research on printed MEs is still important to make it applicable in practice. The two fields of MER research overlap, but there are also two major differences. First, the offline handwritten MER has the extra challenge of touching symbols, which makes it harder to separate them \cite{chan_mathematical_2000, aggarwal_survey_2022}. Second, the characteristics of the benchmark datasets are different. Handwritten (offline) MER uses the CROHME datasets \cite{xie_icdar_2023} as the benchmark, with a vocabulary of $142$ tokens and, on average, $18$ tokens per ME. On the other hand, the printed MER benchmark dataset \emph{im2latex-100k} \cite{kanervisto_im2latex-100k_2016} has a much larger vocabulary of $500$ tokens, which is $3.5$ times greater than the CROHME dataset. Additionally, on average, each ME in the \emph{im2latex-100k} dataset has $2.8$ times as much tokens.\\
However, both MER systems comprise three stages: symbol segmentation, symbol recognition, and 2D structure analysis \cite{aggarwal_survey_2022}. Classic approaches, as the Infty system \cite{suzuki_infty_2003, malon_mathematical_2008} solve these stages separately, whereas end-to-end approaches address them all at once. 
With recent progress in deep learning, end-to-end approaches with an encoder-decoder structure have become prevalent \cite{deng_image--markup_2017}. These systems directly map input images to a semantic text representation, e.g., LaTeX. In general, the encoder is based on convolutional layers to calculate features of the image. The decoder generally uses LSTMs \cite{hochreiter_long_1997}, GRUs \cite{cho_learning_2014}, or Transformers \cite{vaswani_attention_2017}, which translate the feature inputs step-by-step into a token sequence \cite{aggarwal_survey_2022}.\\
\emph{WYGIWYS}, introduced by Deng et \textit{al}. \cite{deng_image--markup_2017}, is one of the first end-to-end MER systems. It calculates its features using a convolutional network stacked with an RNN row encoder. The token sequence is predicted by an RNN decoder with visual attention stacked with a classifier layer. Because of the end-to-end approach, large datasets are required for training. Therefore, the authors introduced \emph{im2latex-100k} \cite{kanervisto_im2latex-100k_2016}, which is still the classic benchmark dataset in printed MER.\\
Cho et \textit{al}. \cite{cho_properties_2014} found that the performance of the encoder-decoder network for text generation declines as the length of the sentence increases. This is particularly relevant for ME sequences, which are usually longer than sentences used in image captioning. As a result, many MER models focus on enhancing the long-distance dependence of the decoder. Various approaches have been developed to overcome this issue.\\
Bian et \textit{al}. \cite{bian_handwritten_2022} developed a bi-directional mutual learning network based on attention aggregation. The network uses two encoders, one that processes the input left-to-right and another that processes it right-to-left. They demonstrated that this structure helps alleviate the issue of long-range dependencies in RNNs.
Li et \textit{al}. \cite{li_when_2022} introduced a method for counting symbols in handwritten MER. Their weakly supervised multi-scale counting module can be combined with most encoder-decoder frameworks, and it improves the model's robustness when the ME is complex and/or long. However, it does not solve the problem with variations in writing styles. 
Yan et \textit{al}. \cite{yan_convmath_2021} developed ConvMath, a printed MER system based entirely on convolutions. They introduced a convolutional decoder to better detect the 2D relation of MEs. 
Markazemy et \textit{al}. \cite{mirkazemy_mathematical_2023} introduced a novel reinforcement learning module to process the decoder output and refine it.\\
Apart from focusing on the decoder, various elements of MER have been researched. Wang et \textit{al}. \cite{wang_image_2019} aimed to enhance the encoder by incorporating DenseNet into printed MER. Li et \textit{al}. \cite{li_improving_2020} introduced scale augmentation and drop attention to handwritten MER to improve the model performance for various ME scales. Peng et \textit{al}. \cite{peng_image_2021} introduced Graph Neural Network in printed MER. However, representing an ME as a graph became popular in handwritten MER \cite{mahdavi_icdar_2019}, but not in printed MER \cite{aggarwal_survey_2022}. Singh \cite{singh_teaching_2018} investigated the visual attention in printed MER and developed two new datasets based on \emph{im2latex-100k}. Furthermore, there have been advancements in the development of end-to-end systems for scientific documents that can recognize not only MEs but also text and tables. However, the current leading end-to-end system from Blecher et \textit{al}. \cite{blecher_nougat_2023} has an Edit distance of only $87.2$\%, which is lower than the best current MER systems.\\
However, to the best of our knowledge, the influence of undesired variations in the GT has not yet been investigated in handwritten nor printed MER.

\section{Detrimental LaTeX Variations}
\label{sec:ProblematicLaTeX}
MEs have a two-dimensional structure which is different from the one-dimensional structure of natural language text. Therefore, a markup language, e.g. LaTeX, is needed to convert MEs into a natural language description. LaTeX is widely used in the scientific community for writing documents. Hence, many MEs in LaTeX exist, making it appealing for printed MER. The widely used benchmark dataset \emph{im2latex-100k} also uses LaTeX to create the MEs for the training and test datasets. However, we discovered two problematic issues with this dataset:
\begin{enumerate}
\item Our analysis of the dataset revealed that the whole \emph{im2latex-100k} dataset was created with a single font. This, on the one hand, drastically limitates the generalisation capability of the performance results reported on this dataset to realistic scenarios where MEs are printed in various different font styles, normally different than the one used for training the systems. This effect was revealed in preliminary experiments when we observed a significant decrease in the performance of all tested systems by changing the font of the test set. This effect is also apparent when we compare the performance results of the baseline models of \emph{im2latex-100k} and \emph{im2latexv2} (refer to Tables \ref{tab:im2latexComparison} and \ref{tab:im2latexv2}).\\
To address this limitation, all MEs of the \emph{im2latex-100k} dataset were generated in many different fonts. 
\item We further discovered another detrimental effect in the GT of \emph{im2latex-100k}: As the GT of the MEs in \emph{im2latex-100k} was taken from real papers written by different authors, there was a large variation of the GT for semantically identical MEs, as illustrated in Figure \ref{fig:LaTeXproblem}. These variations have nothing to do with improved generalization capabilities to be learned or shown. To the contrary: First, it reduces the validity of performance result comparisons of the different systems if this occurs in the test dataset. Second, it is detrimental for the learning of MER systems, if it occurs in the training dataset (by teaching the model that the same input has ambiguous output, leading to reduced learning \cite{simmler_survey_2021}).
In order to minimize these meaningless variations in the GT of \emph{im2latex-100k}, we adopted a data-centric approach to develop a new LaTeX normalization procedure. The data-centric approach involves three steps. 
First, the model is trained using the existing training data. 
Second, the performance of the trained model is evaluated to identify any error patterns. 
Third, these error patterns are utilized to improve the training dataset (in our case by adjusting the LaTeX normalization). 
\end{enumerate}
These steps are repeated until no more error patterns can be detected.
During this iterative process, we have identified six problematic aspects in the GT of the \emph{im2latex-100k} dataset: mathematical fonts, white spaces, curly brackets, sub- and superscript order, tokens, and arrays. These problematic aspects together with our proposed solutions are described in Sections \ref{subsec:mathematicalFonts} - \ref{subsec:arrays}. We designed our normalization process to address these issues and reduce undesired variations. 
\subsection{Mathematical Fonts}
\label{subsec:mathematicalFonts}
\begin{figure}
    \centering
    \includegraphics[width=.4\textwidth]{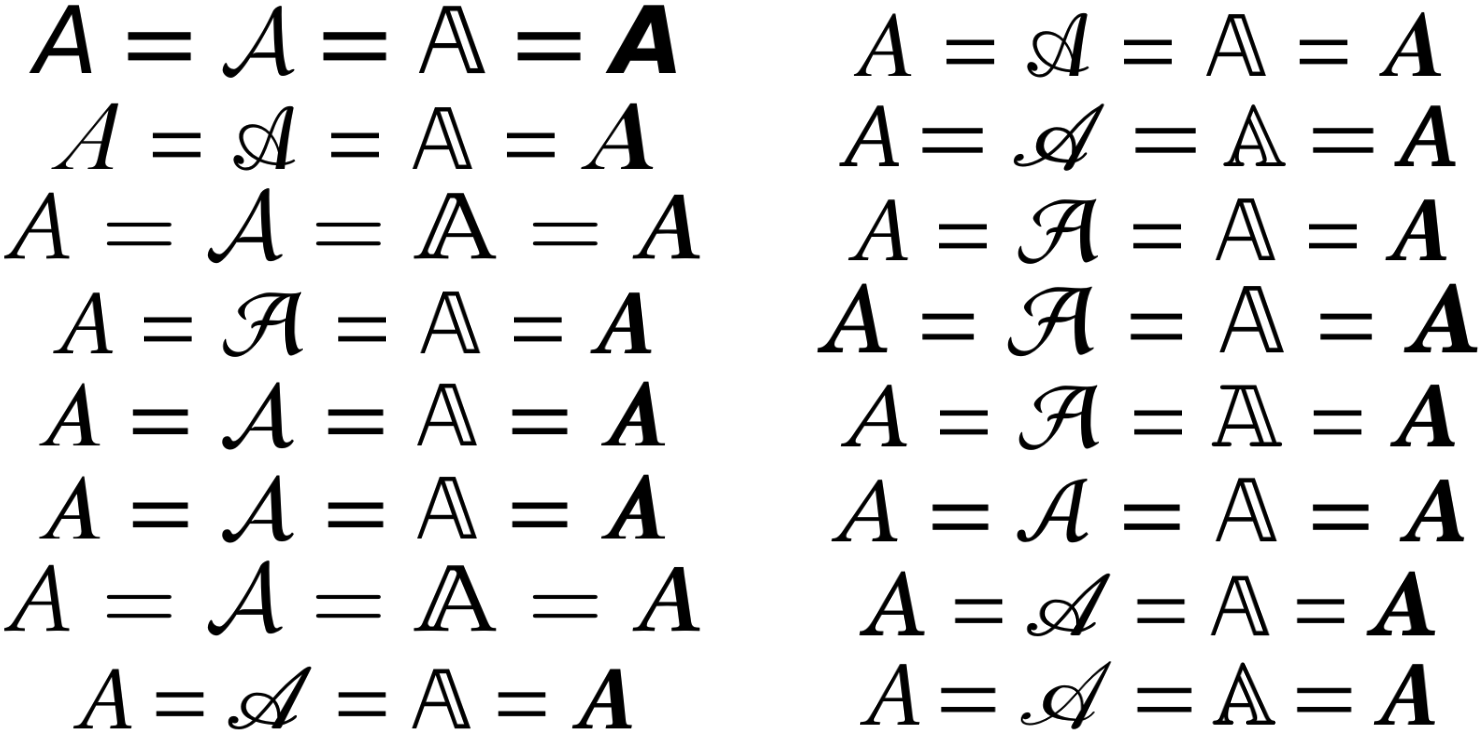}
    \caption{The ME: A = \textbackslash mathcal\{A\} = \textbackslash mathbb\{A\} = \textbackslash boldsymbol\{A\} generated with the $16$ fonts, which can render all three basic mathematical fonts.}
    \label{fig:styleTypes}
\end{figure}

Using different mathematical fonts, such as bold or double bold, to indicate vectors or spaces can be challenging for MER. Recognizing these mathematical fonts is simple if only one font is used for all MEs, but it becomes challenging with multiple fonts, as shown in Figure \ref{fig:styleTypes}. Additionally, it can be challenging to create a dataset with mathematical fonts, as not all mathematical font commands work with every font, i.e., only $16$ out of $59$ fonts respond to the three basic mathematical font commands (\verb|\mathcal|, \verb|\mathbb|, and \verb|\boldsymbol|) for all symbols. As a result, the collected ME can contain a mathematical font command that does not influence the compiled image of the ME. To avoid this, we decided to remove all mathematical font commands, which is a simplification of the task but reduces the number of labeling errors in the GT.

\subsection{White Spaces}
In LaTeX, authors can adjust the white space between two symbols using various commands (e.g., \verb|\quad|). However, these commands are primarily defined relative to the font size, making it essential for the model to accurately detect the font size, which is influenced by the font. 
Additionally, multiple combinations of white space commands exist for each relative white space length. This makes it impossible for the model to predict the white space commands when multiple fonts are utilized, and the white space commands do not follow a clear pattern. Since the white space does not impact the canonical form of an ME, we decided to remove all white space commands from the GT.
\subsection{Curly Brackets}
In LaTeX, curly brackets are used to define the scope of LaTeX commands. As a result, $33$\% of all tokens in \emph{im2latex-100k} are curly brackets. However, the issue with curly brackets is that they are often optional and can be added without changing the visual appearance of a mathematical expression (e.g., \verb|a_3|, \verb|a_{3}|, and \verb|{a_{3}}| are visually identical). Therefore, we introduced a precise definition of which curly brackets are required and which are not. This reduces ambiguity and the number of curly brackets in the GT.
\subsection{Sub- and Superscript Order}
Symbols can have sub- and superscripts but the order in LaTeX code is irrelevant for the visual appearance of the ME. If multiple sub- and superscripts (e.g. \verb|a^{b}_{c}^{d}|) exist, we decided to combine these to one subscript and one superscript (e.g. \verb|a_{c}^{bd}|) to reduce ambiguity and the number of tokens. Although this normalization steps may result in errors in certain circumstances, it typically minimizes undesired variations of the GT.
\subsection{Tokens}
We identified three issues on the token level.
First, many expressions in LaTeX exist that produce the same visual symbol (e.g., \verb|\ge| to \verb|\geq|). Hence, we identified all redundant LaTeX expressions in \emph{im2latex-100k} and replaced them by the canonical form. 
Second, some tokens in the ME imply that the ME not only contains mathematical elements (e.g., \verb|\cite|, \verb|\label|) or is more a graphic element than an ME (e.g., \verb|\fbox|). Hence, we decided to delete all MEs with such tokens.
Third, the tokenizer introduced by Deng et \textit{al}. \cite{deng_image--markup_2017} sometimes combines two LaTeX commands in one token, e.g., the token \verb|\right{| actually contains two tokens \verb|\right| and \verb|{|. This can increase the vocabulary and introduce undesired variation. To avoid this, we split up these tokens, so each token represents only one LaTeX command.
\subsection{Arrays}
\label{subsec:arrays}
The array structure has the purpose to arrange elements in a grid, e.g., a matrix. However, many authors use this feature to align MEs instead (e.g., \verb|\begin{array}{cc}a=b,&c=d\end{array}|). Additionally, the column alignment indicators (l, c, r) do not affect the semantics of the array. Moreover, not all arrays are well-defined and may contain empty columns, rows, or cells. Hence, we removed array structures used only for formatting, and reduced GT variation in the array structures by replacing all column indicators with c. We also removed sparse arrays (empty entries or number of columns doesn't match number of column alignment indicators).

\section{Applying LaTeX Normalization for Printed MER}
\label{sec:ApplyingLaTeXNormalizationforPrintedMER}
To evaluate our LaTeX normalization, we applied it for printed MER. Therefore, we developed an enhanced version of \emph{im2latex-100k} described in Section \ref{subsec:im2latexv2}, a real-world test set described in Section \ref{subsec:realFormula}, and a new printed MER model described in Section \ref{subsec:mathnet}. Lastly, Section \ref{subsec:metrics} gives an introduction to printed MER metrics.\\
\subsection{im2latexv2}
\label{subsec:im2latexv2}
\begin{figure}
    \centering
    \includegraphics[width=.4\textwidth]{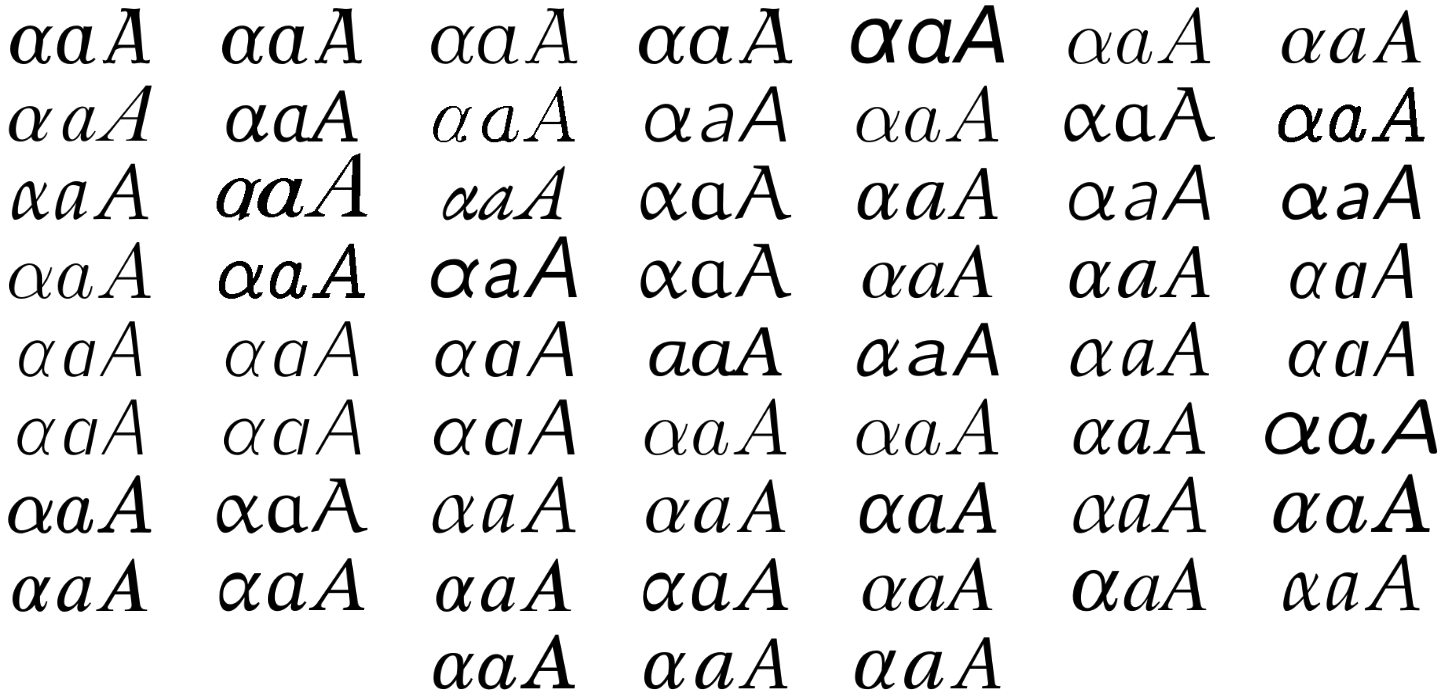}
    \caption{Overview of all 59 fonts in the im2latexv2 dataset.}
    \label{fig:allStyles}
\end{figure}
This dataset is an evolution of \emph{im2latex-100k} and contains three major modifications over existing printed MER datasets. 
First, we used the normalization procedure described in Section \ref{sec:ProblematicLaTeX} with minor modifications for rendering. To create controlled visual diversity, we left the column alignment indicators of arrays unchanged and did not remove the \verb|\right| and \verb|\left| tokens for rendering the MEs. Using the normalized MEs we can ensure that the GT and image coincide.
In comparison, Deng et \textit{al}. \cite{deng_image--markup_2017} used the original ME descriptions for \emph{im2latex-100k}. Hence, the GT for the same image may vary.\\
Second, in contrast to \emph{im2latex-100k}, \emph{im2latex-90k}, and \emph{im2latex-140k}, we rendered each ME with $30$ different fonts for the training dataset and $59$ for the validation and test set. The incorporation of multiple fonts makes the dataset more realistic. Furthermore, $29$ fonts only appear in the validation and test set to assess a model's generalization capability. The font variation introduced this way is illustrated in Figure \ref{fig:allStyles}.\\
Third, we used $600$ DPI (font size $12$pt) to render the images, because down-sampling works well compared to up-sampling. In contrast, Deng et \textit{al}. \cite{deng_image--markup_2017} suggested $100$ DPI for the MER task. Singh \cite{singh_teaching_2018} used $200$ DPI and in handwritten MER different resolutions exist. However, the scanned images of handwritten MER correspond to resolutions of $300$ to $600$ DPI in printed MER. We will demonstrate the influence of the resolution on the model performance in Section \ref{subsec:optimalResolution}.\\
The resulting \emph{im2latexv2} dataset contains fewer MEs than the original \emph{im2latex-100k} due to our rendering pipeline, which includes four check criteria (see Algorithm \ref{alg:latexNormalizationProcess} and Table \ref{tab:im2latexv2Overview}). $19$ MEs in the training set and $30$ MEs in the test set had to be dropped because the image was blank. Additionally, we found 1 empty ME in the train set and 42 empty MEs in the test set. We manually corrected the empty ME in the train set and $37$ MEs in the test set. We removed the other $5$ empty MEs from the test set because the image depicted a drawing rather than a valid ME. Besides, our normalization step dropped $882$ MEs in the training set, $116$ in the validation set, and $179$ MEs in the test set. The rendering step removed $129$ MEs in the training set, $11$ MEs in the validation set, and $23$ MEs in the test set, which could not be rendered for all fonts. As a result, the training set was reduced by $1'023$ MEs, the validation set by $127$ MEs, and the test set by $237$ MEs compared to the original \emph{im2latex-100k}. The new normalized dataset \emph{im2latexv2} finally contains approximately $92'600$ MEs (ref. Table \ref{tab:im2latexv2Overview}).
\begin{table*}
    \caption{Overview of the reasons, why we deleted different MEs.}
    \label{tab:im2latexv2Overview}
    \centering
    \begin{tabular}{c|c|c|c|c}
        Category & Alg. \ref{alg:latexNormalizationProcess} line & train & validation & test \\
        \hline
        im2latex-100k & & $75'275$ & $8'370$ & $10'355$\\
        \hline
        white image & $2$ & $19$ & $0$ & $30$\\
        empty ME (corrected) & $5$ & $1$ ($1$) & $0$ & $42$ ($37$)\\
        normalization step & $10$ & $882$ & $116$ & $179$\\
        rendering errors & $15$ & $129$ & $11$ & $23$\\
        \hline
        im2latexv2 & & $74'245$ & $8'243$ & $10'118$\\
    \end{tabular}
\end{table*}

\begin{algorithm}
\caption{Im2latexV2 Rendering Pipeline}\label{alg:latexNormalizationProcess}
\begin{algorithmic}[1]
\Require{$[F_{1}, I_{1}] \dots [F_{N}, I_{N}]$ in im2latex-100k}
\For{$k \gets 1$ to $N$}
    \If{$I_k$ is $null$}
        \State{\textbf{continue}}
    \EndIf
    \If{$F_k$ is $null$}
        \State{check manually}
        \State{\textbf{continue}}
    \EndIf
    \State{$F_k \gets $LaTeXNormalization($F_k$)}
    \If{$F_k$ is $null$}
        \State{\textbf{continue}}
    \EndIf
    \For{$j, r$ in $renderingSetups$}
        \State{$I_{k,j}, e_{k,j} \gets $renderImages($F_k, renderingSetup$)}
        \If{$e_{k, j}$ is not $null$}
            \State{\textbf{break}}
        \EndIf
    \EndFor
\EndFor
\end{algorithmic}
\end{algorithm}

\subsection{realFormula}
\label{subsec:realFormula}
By using the Mathematical Formula Detection model from Schmitt-Koopmann et \textit{al}. \cite{schmitt-koopmann_formulanet_2022}, we collected over $250$k ME from randomly selected arXiv papers with $600$ DPI and selected $200$ MEs at random for manual annotation. As shown in Table \ref{tab:realFormulaIssues} we deleted $69$ MEs where the image was larger than $768$x$2400$ pixels. Nine other MEs were deleted because the image did not show the complete ME and $1$ ME showed a sparse matrix. Hence, we manually annotated $121$ MEs. Of these $121$ MEs, $110$ were single-line MEs and $11$ were multi-line MEs. Five single-line MEs contained an array, and $43$ MEs contained style types (\verb|\boldsymbol|, \verb|\mathbb|, \verb|\mathcal|).
\begin{table}
    \caption{Overview of $200$ randomly selected MEs. It shows various issues that arose, requiring some MEs to be excluded from the \emph{realFormula} set.}
    \label{tab:realFormulaIssues}
    \centering
    \begin{tabular}{c|c}
        Category & Number of images \\
        \hline
        too large & $69$\\
        cropping error & $9$\\
        sparse matrix & $1$\\
        \hline
        removed & $79$\\
        correct & $121$\\
    \end{tabular}
\end{table}
\subsection{MathNet}
\label{subsec:mathnet}
\begin{figure*}
    \centering
    \includegraphics[width=0.8\linewidth]{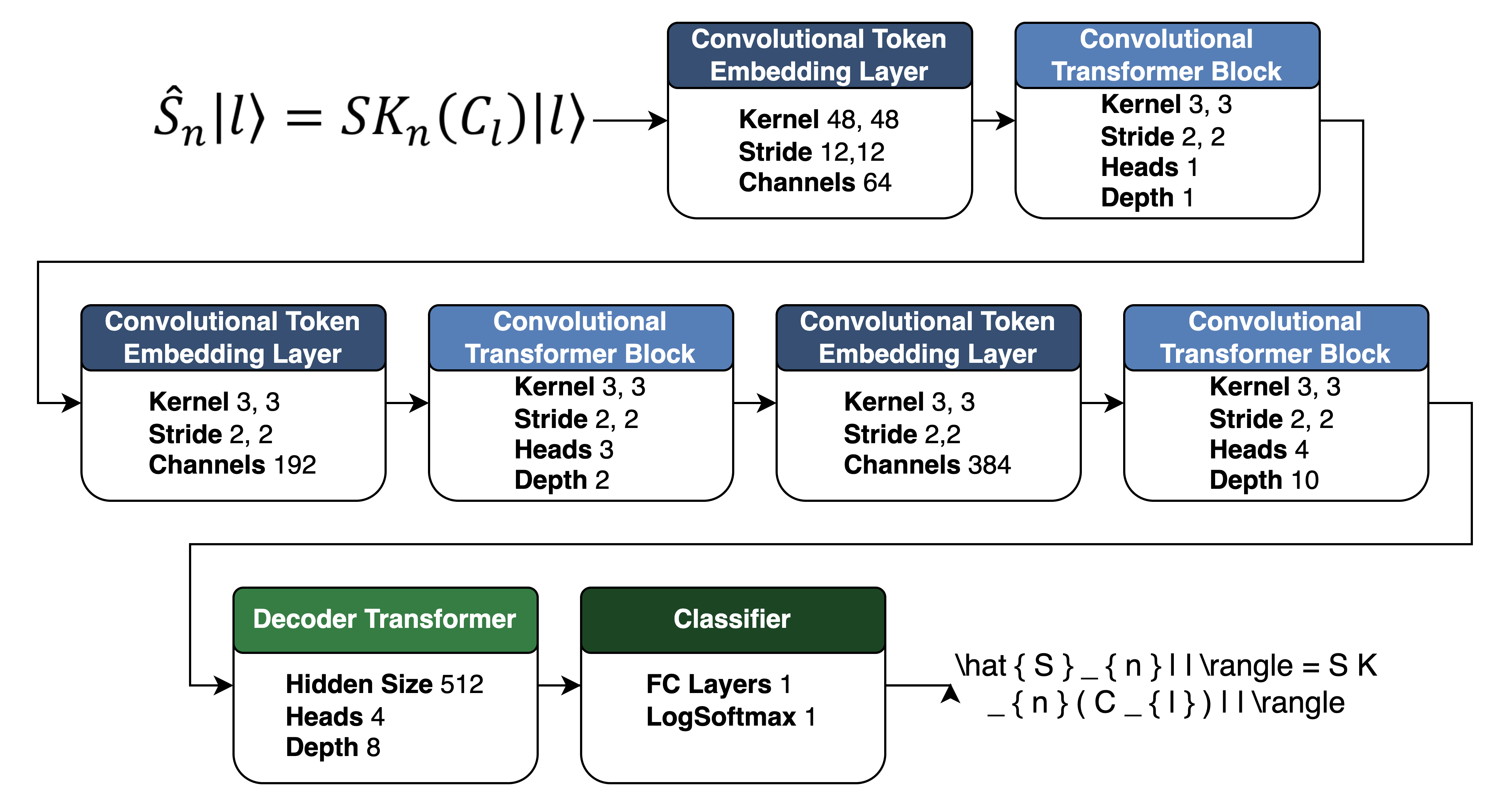}
    \caption{Overview of our MER model, called \emph{MathNet}. The CvT consists of $3$ layers, which are a combination of an embedding layer and a transformer block. The encoded image is decoded with a decoder transformer and a classifier layer.}
    \label{fig:modelOverview}
\end{figure*}

For our experiments we decided to use an encoder-decoder approach similar to the state of the art MER models.\\
In order to accurately process ME images, it is crucial for the encoder to extract informative features. This requires the encoder to be able to focus on small details while also considering the overall structure of the ME, such as a fraction. To handle both short-term and long-term relationships, Deng et \textit{al}. \cite{deng_image--markup_2017} developed the Coarse-to-Fine Attention mechanism. However, recent advancements in image recognition have shown that vision transformers (ViTs) \cite{dosovitskiy_image_2021} are well-suited for this task. A further development of ViTs are convolutional vision transformers (CvTs) \cite{wu_cvt_2021}. CvTs combine convolutions with transformers, resulting in superior performance and efficiency with a smaller model. Hence, we decided to use a CvT instead of a usual CNN encoder.\\
The decoder is responsible for converting the features of the encoder into the chosen markup language, i.e. LaTeX. Unlike most other MER systems, \emph{MathNet} uses a regular decoder transformer instead of LSTMs. Vaswani et \textit{al}. \cite{vaswani_attention_2017} showed that transformers are better suited for handling long sequences, as we have in printed MER. Furthermore, \emph{im2latexv2} is much larger than \emph{im2latex-100k}, which should benefit the training of transformers. Our decoder transformer has $8$ heads and a depth of $4$. On top of this, we added a classifier layer with a log softmax. An overview of our \emph{MathNet} model with the layer sizes is shown in Fig. \ref{fig:modelOverview}.\\
We used a cross-entropy loss between the GT sequence and the predicted sequence. To optimize our model, we used the Adam optimizer \cite{kingma_adam_2014} with an initial  learning rate of $0.000075$ and a batch size of $36$. Our model was trained on a single Nvidia Tesla V100-SXM2-32GB GPU.

In order to prevent the model from learning useless patterns in images, we applied different augmentation techniques. We have identified four patterns that the model should explicitly not learn. Firstly, to avoid confusion by white spaces (such as \verb|\quad| and \verb|\,|), we randomly introduced white pixel columns to the image. Secondly, we used blurring masks and changed the image resolution randomly. Thirdly, we resized the image to prevent the model from focusing on a specific text size. Lastly, we added a white border of variable size to the image to facilitate batch-wise processing and ensure that all images have the same size.

\subsection{Metrics}
\label{subsec:metrics}
Printed MER primarily uses three metrics (Edit distance, Bleu-Score, and Exact Match) to evaluate model performance. 
An edit distance counts the operations needed to transform one sequence into another sequence. Depending on the operations allowed, different edit distances exist. The most popular edit distance for printed MER and the one we use is the Levenshtein edit distance ($lev$). It contains three operations: 1) insert a new token, 2) delete a token, and 3) replace a token. The Edit score, as used by Deng et \textit{al}., is the edit distance normalized by the max sequence length of the GT and predicted sequence (PRE) as shown in Eq. \ref{eq:editScore}. For MER, an Edit score of $100\%$ is a perfect prediction.
\begin{equation}
    Edit Score = (1 - \frac{lev(GT, PRE)}{max(len(GT), len(PRE))}) \cdot 100\%
    \label{eq:editScore}
\end{equation}
The Bleu score compares subsequences of two sequences. A predefined number, usually 4 in MER, determines the maximum subsequence length. To determine the Bleu score we create n-grams for the sequences and then calculate the precision between the n-grams of sequences. The Bleu score is the average precision with a brevity penalty to discourage overly short predictions. However, the Bleu score is designed for longer sequences, and errors at the sequence borders count less than errors in the middle. This behavior can significantly impact the score of MEs.
Exact Match measures the amount of fully correct MEs. It makes no distinction between partially correct ME and completely incorrect ME.\\
With respect to PDF accessibility, we regard the Edit score as the most relevant metric because it indicates the amount of work that must be manually done to correct all errors in a recognized ME. The Bleu-4 score shows unwanted behavior for short MEs and focuses more on patterns than on the correct order of the tokens. The interpretation of exact match is very limited through the binary output.\\
However, as discussed in Section \ref{sec:ProblematicLaTeX}, semantically identical MEs can be produced with different LaTeX code sequences. Hence, the metric results are largely influenced by the used tokenizer and normalization. For example, if the $x$ in Eq. \ref{eq:EditDistanceExample} should be a $5$, the Edit score would rise from $96.3$\% (1 of 27) to $97.8$\% (1 of 45) if curly brackets had been added in the GT around each entry in the array, which would mean a $40.5$\% reduction of the Edit error rate (1 - Edit score). The Bleu-4 score shows similar behavior.\\ 
\begin{equation}
\left ( \begin{array} { c c c }
1 & 2 & 3\\
4 & x & 6\\
7 & 8 & 9\\
\end{array} \right )
\label{eq:EditDistanceExample}
\end{equation}

\section{Experiments with Printed MER}
\label{sec:experiments}
This section presents the results of four experiments. The first experiments (see Section \ref{subsec:optimalResolution}) demonstrate the influence of the image resolution on the model performance. Experiments two to five are comparison experiments with printed MER models. To ensure a fair comparison, we used the provided pre-trained models (\emph{WYGIWYS}, \emph{i2l-strips}, \emph{i2l-nopool}) and normalized the predictions with our normalization process.\\ 
We used the following four datasets to compare the models: 1) The benchmark dataset \emph{im2latex-100k}. 2) Our enhanced version \emph{im2latev2} which includes multiple fonts. 3) Our developed real-world dataset \emph{realFormula} with MEs images extracted from papers, to demonstrate how well the systems perform in a real-world environment. 4) \emph{InftyMDB-1} contains MEs images, which were scanned with $600$ DPI. This dataset is also intended to evaluate real-world performance, specifically the impact of scanning noise.\\
It is important to note that \emph{i2l-strips} and \emph{i2l-nopool} used a modified dataset built upon \emph{im2latex-100k} which had a different split between training, validation, and testing. Hence, MEs of the test sets of \emph{im2latex-100k} and \emph{im2latexv2} could be in the training set of \emph{i2l-strips} and \emph{i2l-nopool}.
\subsection{Optimal Image Resolution}
\label{subsec:optimalResolution}
When the resolution is low, the image has fewer details and the model has to focus on the general structure. On the other hand, high-resolution images provide more detail which can enable the model to differentiate better between symbols.\\
However, there is no clear definition what the optimal image resolution for MER is (with the standard font size of $12$pt). According to Deng et \textit{al}. \cite{deng_image--markup_2017}, $100$ DPI images are recommended, while Singh \cite{singh_teaching_2018} used images with $200$ DPI. In contrast, handwritten MER mainly uses image sizes that correspond to resolutions between $300$ and $600$ DPI \cite{xie_icdar_2023}.\\
We trained our model on various image resolutions to demonstrate the impact of this resolution, as shown in Table \ref{tab:optimalResolution}. We used $100$, $200$, $300$, and $600$ DPI image resolutions. The images of the test set were scaled accordingly. Our results reveal a significant improvement between $100$ and $200$ DPI. Moreover, the model's performance still improves with even higher resolutions. However, we did not test resolutions higher than $600$ DPI because $600$ DPI is typically the maximum for scanned documents. For the subsequent experiments, we used the model with $600$ DPI training images.\\

\begin{table}
    \caption{Influence of the training image resolution on the \emph{im2latexv2} test set. We trained our model with $100$, $200$, $300$, and $600$ DPI.}
    \label{tab:optimalResolution}
    \centering
    \begin{tabular}{c|c|c|c}
        DPI & Edit [\%] & Bleu-4 [\%] & EM [\%]\\
        \hline
        $100$ & $78.2$ & $66.0$ & $6.0$\\
        $200$ & $93.5$ & $93.2$ & $56.9$\\
        $300$ & $96.9$ & $96.9$ & $75.0$\\
        $600$ & $\boldsymbol{98.0}$ & $\boldsymbol{98.2}$ & $\boldsymbol{84.9}$\\
    \end{tabular}
\end{table}

\subsection{im2latex-100k}
\label{subsec:resultsIm2latex}
This Section presents the results on the \emph{im2latex-100k} test set. \emph{im2latex-100k} contains images with $100$ DPI. However, \emph{i2l-strips} and \emph{i2l-pool} were trained with $200$ DPI images, and our model was trained with $600$ DPI images so they require larger images. We used two techniques to create larger images. First, we resized the original images with OpenCV to the training size. Second, we rendered the original MEs without normalization using a LaTeX environment to eliminate the influence of insufficient resolution.\\
As shown in Table \ref{tab:im2latexComparison}, rendering an image with a higher resolution achieves better results as resizing the original images. \emph{MathNet} achieved the same Edit scores ($88.6$\%) as \emph{WYGIWYS} with the resized images. However, the Edit error rate (1 - Edit score) nearly halved from $11.4$\% to $5.3$\% when the images were rendered with $600$DPI. \emph{i2l-strips} and \emph{i2l-nopool} performed poorly with the resized images ($32.5$\% and $32.0$\%), but similarly to \emph{WYGIWYS} with the rendered images ($86.9$\% and $86.8$\%). Interestingly, the exact match score of \emph{MathNet} is low compared to the other systems. Hence, the fewer errors of \emph{MathNet} must be more widely spread over the different MEs than those of the other systems.

\begin{table}
    \caption{Results of the \emph{im2latex-100k} test set. We run the models once with the original images, resized to the training size, and once with the images rendered with the optimal resolution.}
    \label{tab:im2latexComparison}
    \centering
    \begin{tabular}{c|c|c|c}
        Model & Edit [\%]& Bleu-4 [\%] & EM [\%]\\
        \hline
        WYGIWYS & $88.6$ & $90.3$ & $\boldsymbol{78.6}$\\
        I2l-strips (resized) & $32.5$ & $12.7$ & $0$\\
        I2l-strips (rendered) & $86.9$ & $86.1$ & $76.3$\\
        I2l-nopool (resized) & $32.0$ & $13.4$ & $0$\\
        I2l-nopool (rendered) & $86.8$ & $85.9$ & $76.2$\\
        MathNet (our) (resized) & $88.6$ & $86.0$ & $31.8$\\
        MathNet (our) (rendered) & $\boldsymbol{94.7}$ & $\boldsymbol{94.5}$ & $63.4$\\
    \end{tabular}
\end{table}

\subsection{im2latexv2}
\label{subsec:resultsIm2latexv2}
This section presents the results with the \emph{im2latexv2} test set. We assigned a random font for each ME in the test set. We used the same font-ME combination for all models to avoid influencing the results by using different fonts for the same ME. Since \emph{im2latexv2} uses $600$ DPI images, we resized the images for \emph{i2l-strips}, \emph{i2l-nopool}, and \emph{WYGIWYS} to the training image resolution. As presented in Table \ref{tab:im2latexv2}, \emph{WYGIWYS}'s Edit score drops dramatically from $88.6$\% to $37.2$\% compared to \emph{im2latex-100k}. However, \emph{i2l-strips} and \emph{i2l-nopool} handle multiple fonts better, with only a small decrease of $11$ pp. and $10.8$ pp. in the Edit score. In contrast, our model shows a $2.5$ pp. increase in the Edit score. We attribute this increase to the fact that \emph{im2latex-100k} includes problematic mathematical fonts, as explained in Section \ref{subsec:styleTypesIssue}.

\begin{table*}
    \caption{Results of the im2latexv2 test set. We resized the images to the optimal size. Errors is the summed Levenshtein Distance over all MEs. Array Errors is the summed Levenshtein Distance of all MEs with an array structure. nA is the Edit Score of all MEs without an array structure.}
    \centering
    \begin{tabular}{c|c|c|c|c|c|c|c}
        Model & Train Dataset & Edit [\%] & Bleu-4 [\%]& EM\ [\%] & Errors & Array Errors & Edit nA [\%]\\
        \hline
        WYGIWYS & im2latex-100k & $37.2$ & $23.9$ & $0$ & $564'700$ & $31'742$ & $37.1$\\
        I2l-strips & im2latex-140k & $75.9$ & $65.9$ & $10.3$ & $143'802$ & $28'539$ & $79.2$\\
        I2l-nopool & im2latex-140k & $76.0$ & $66.4$ & $10.4$ & $144'860$ & $28'015$ & $78.9$\\
        \textbf{MathNet (our)} & im2latexv2 & $\boldsymbol{97.2}$ & $\boldsymbol{96.8}$ & $\boldsymbol{83.9}$ & $\boldsymbol{16'596}$ & $\boldsymbol{8'728}$ & $\boldsymbol{98.6}$\\
        \hline
        \textbf{MathNet (our)} & im2latex-100k & $78.2$ & $65.9$ & $10.3$ & & &\\
        \textbf{MathNet (our)} & im2latexv2 (vanilla font) & $90.4$ & $84.3$ & $26.4$ & & &\\
    \end{tabular}
    \label{tab:im2latexv2}
\end{table*}

\subsection{realFormula}
\label{subsec:resultsRealFormula}
This section presents the results of the \emph{realFormula} test set. Table \ref{tab:realFormulaComparison} provides an overview of the results. The table shows that our model reaches an Edit score of $88.3$\%. This is about three times higher than \emph{WYGIWYS} ($27.5$\%) and approximately one-third higher than \emph{i2l-strips} ($65.1$\%) and \emph{i2l-nopool} ($65.2$\%). In order to quantify the impact of multi-line formulae  we have split the MEs into multi-line (M) and single-line (S) MEs as discussed in Section \ref{subsec:multilineME}. To determine the influence of the array element we have filtered out all MEs with the token elements \verb|\begin{array}| and \verb|\end{array}| (nA), which is discussed in Section \ref{subsec:arrayIssue}. Additionally, we have filtered out all MEs with mathematical fonts (nMF); this issue is discussed in Section \ref{subsec:styleTypesIssue}.

\begin{table*}
    \caption{Edit scores [\%] of the \emph{realFormula} test set. S: single line ME, M: multi-line ME, nA: no arrays, A: arrays, nMF: no mathematical fonts, MF: mathematical fonts}
    \centering
    \begin{tabular}{c|c|c|c|c|c|c|c}
        Model & all & S & S nA & S A & S nMF & S MF & M \\
        \hline
        WYGIWYS & $27.5$ & $28.6$ & $29.6$ & $11.6$ & $28.9$ & $27.8$ & $22.5$ \\
        I2l-strips & $65.1$ & $76.6$ & $81.7$ & $17.7$ & $82.6$ & $65.1$ & $15.9$ \\
        I2l-nopool & $65.2$ & $77.1$ & $81.9$ & $20.7$ & $83.5$ & $65.0$ & $13.9$ \\
        MathNet (our) & $\boldsymbol{88.3}$ & $\boldsymbol{92.5}$ & $\boldsymbol{93.3}$ & $\boldsymbol{84.1}$ & $\boldsymbol{94.1}$ & $\boldsymbol{89.5}$ & $\boldsymbol{71.2}$ \\
    \end{tabular}
    \label{tab:realFormulaComparison}
\end{table*}

\subsection{InftyMDB-1}
\label{subsec:resultsInftyMDB1}
This Section presents the results on the \emph{InftyMDB-1} test set \cite{fujiyoshi_inftymdb-1_2009}. \emph{InftyMDB-1} contains $4400$ images of scanned MEs with a resolution of $600$ DPI. We used the pandoc library to covert the MathML GT into LaTeX GT and processed the resulting LaTeX strings similar to the other datasets.\\
As shown in Table \ref{tab:InftyMDB1}, the resulting performance of \emph{MathNet} is about the same compared to \emph{realFormula} test set. However, it demonstrates that \emph{MathNet} is not significantly affected by the noise of the scanning process. In contrast, the performance of \emph{WYGIWYS}, \emph{i2l-strips}, and \emph{i2l-nopool} drops by $10.2$ pp., $2.0$ pp., and $1.6$ pp.. This highlights that these models are probably affected by the noise of the scanning process. However, since our focus is on scientific PDFs, we assume that scientific PDFs are usually available in native digital format. Hence, scanned documents with geometric deformation, coloration, and noise are considered as not in our research focus.

\begin{table}
    \caption{Prediction results on the InftyMDB-1 dataset ( scanned MEs).}
    \centering
    \begin{tabular}{c|c|c|c}
        Model & Edit [\%]& Bleu-4 [\%] & EM [\%]\\
        \hline
        WYGIWYS & $17.3$ & $7.1$ & $0$\\
        I2l-strips & $63.5$ & $46.5$ & $4.1$ \\
        I2l-nopool & $63.2$ & $46.4$ & $3.4$\\
        MathNet (our) & $\boldsymbol{89.2}$ & $\boldsymbol{85.4}$ & $\boldsymbol{35.4}$ \\
    \end{tabular}
    \label{tab:InftyMDB1}
\end{table}

\section{Discussion}
\label{sec:discussion}

\subsection{Data Related Achievements and Challenges}
\label{subsec:DataRealtedDiscussion}
As our results reveal, our data-centric approach with the LaTeX normalization and augmentation process is very beneficial for the training of robust printed MER models. The influence of our normalization and the use of multiple fonts on the model performance is discussed in Section \ref{subsec:ImpactIm2latexv2}. Section \ref{susec:newFonts} demonstrates that our model is adept at working with fonts not included in the training set. \\
However, in our error analysis, we encountered two significant challenges. First, the array element was the main culprit of errors, as detailed in Section \ref{subsec:arrayIssue}. Second, the absence of mathematical fonts and multi-line MEs in the \emph{im2latexv2} training dataset poses a challenge for our model on the \emph{realFormula} test set, as discussed in Section \ref{subsec:styleTypesIssue} and Section \ref{subsec:multilineME}. Section \ref{subsec:problematicTokens} gives an overview of the most frequent token errors with MathNet and \emph{im2latexv2}. 

\subsubsection{The Impact of Normalization and Multiple Fonts}
\label{subsec:ImpactIm2latexv2}
We conducted experiments to separately analyse the influences of our model architecture, our normalization process, and the use of multiple fonts. We trained the model three times, once with the \emph{im2latex-100k} dataset, once with the \emph{im2latexv2} dataset using only the basic font, and once with the full \emph{im2latexv2} dataset. The results are shown in Table \ref{tab:im2latexv2}.\\
When we used the \emph{im2latex-100k} dataset, our model's Edit score ($78.2$\%) was more than double that of \emph{WYGIWYS} ($37.2$\%) and was $2.3$ pp.  $2.2$ pp. higher than \emph{i2l-strips} and \emph{i2l-nopool}. This demonstrates the beneficial network design of our model. The advantage of our model architecture is analyzed further in Section \ref{subsec:modelRelatedDiscussion}. However, the normalization process has a much stronger impact on the model's Edit score, with a $12.2$ pp. improvement when using the \emph{im2latexv2} dataset with the vanilla font for all MEs. The remaining $6.8$ pp. improvement is explained by the use of multiple fonts for the MEs during training. In summary, the model architecture is marginally better as state of the art model architectures. However, two-thirds of the improvement compared to state of the art models are due to our LaTeX normalization process, while the remaining third is attributed to the use of multiple fonts. This reveals the significant influence of our LaTeX normalization process during model training and, hence, the value of the new dataset \emph{im2latexv2}.\\

\subsubsection{Non-Training Fonts}
\label{susec:newFonts}
The \emph{im2latexv2} training set only includes 30 of the 59 fonts in the test set. We tested our model's ability to work with fonts not in the training set and found that that the font influence is negligible. Specifically, we achieved a $97.5$\% Edit score for MEs with fonts in the training set and $96.8$\% for MEs with fonts that were not. Overall, this demonstrates our model's strong generalizability.

\subsubsection{Array Issue}
\label{subsec:arrayIssue}
LaTeX users normally use the array structure (\verb|\begin{array}| $\cdots$ \verb|\end{array}|) to create a matrix, but some authors use it to format their ME instead. To address this unwanted variation we introduced normalization steps in Section \ref{sec:ProblematicLaTeX} to reduce the use of the array environment for formatting purposes. However, even with our normalization process, the array structure remains challenging for MER, as shown in Table \ref{tab:im2latexv2}. Out of all the prediction errors on the \emph{im2latexv2} test set with our model, $52.6$\% are related to MEs with an array structure. However, this array structure is only present in $4.8$\% of all the MEs. Therefore, by removing MEs that use the array structure, our model's Edit error rate is reduced by $50$\% (from $2.8$\% to $1.4$\%). \emph{i2l-strips} and \emph{i2l-nopool} also see reductions from $24.1$\% to $20.8$\% and from $24$\% to $21.1$\%, respectively. The effect on \emph{WYGIWYS} is not significant. We attribute this to the high overall Edit error rate of \emph{WYGIWYS}. The same problems with array structures can be seen in the results of Table \ref{tab:realFormulaComparison} for the \emph{realFormula} test set. For instance, \emph{MathNet} achieves an Edit score of $93.3$\% for single line MEs without arrays and $84.1$\% for single line MEs with arrays.

\subsubsection{Mathematical Font Issue}
\label{subsec:styleTypesIssue}
In ME, changes to the font style of symbols (mathematical fonts) are used to indicate, e.g., vectors and spaces. As these mathematical fonts are not rendered correctly for all fonts, we decided to remove all mathematical font tokens in \emph{im2latev2} to ensure the images are rendered correctly. Consequently, \emph{MathNet} cannot detect mathematical fonts. Our results with the \emph{realFormula} test set reveal that removing MEs with mathematical fonts in the training set has a significant influence on the model's real-world performance. Without counting the mathematical font tokens as an error, the Edit score of MEs without mathematical fonts is $94.1$\% (column S nMF), whereas MEs rendered with mathematical fonts, it drops to only $89.5$\% (column S MF). Table \ref{tab:styleTypes} shows that MEs rendered with the three mathematical fonts \verb|\mathcal|, \verb|\mathbb|, and \verb|\operatorname| are especially challenging for \emph{MathNet}. In contrast, the mathematical font \verb|\boldsymbol| has no negative influence on the performance. Nevertheless, mathematical fonts are a limitation of \emph{MathNet} and \emph{im2latexv2} and, hence, the predicting results of \emph{MathNet} deteriorate for MEs containing mathematical fonts. This issue is to be addressed in future research.

\begin{table}
    \caption{Prediction results for MEs in the \emph{realFormula} test set with mathematical fonts.}
    \label{tab:styleTypes}
    \centering
    \begin{tabular}{c|c|c|c}
        Mathematical Font Type & Edit [\%] & Bleu-4  [\%]& EM [\%]\\
        \hline
        \verb|\boldsymbol| & $98.4$ & $96.4$ & $63.6$\\
        \verb|\mathcal| & $90.4$ & $83.7$ & $18.2$\\
        \verb|\mathbb| & $84.6$ & $75.4$ & $0.0$\\
        \verb|\operatorname| & $83.4$ & $75.7$ & $0.0$\\
    \end{tabular}
\end{table}

\subsubsection{Multi-line ME}
\label{subsec:multilineME}
The MEs in the \emph{im2latex-100k} dataset are limited to $150$ tokens, so there are almost no multi-line MEs. However, in the \emph{realFormula} dataset
we had to drop $69$ MEs because they were too large, and these were all multi-line MEs. Together with the $11$ multi-line MEs in the final dataset, $80$ of the original $200$ MEs were multi-line MEs (see Table \ref{tab:realFormulaIssues}). This reveals that a real-world MER needs to handle multi-line MEs in addition to single-line MEs.\\
Table \ref{tab:realFormulaComparison} shows that our model performs $30$\% better in terms of Edit score for single-line MEs ($92.5$\%) compared to multi-line MEs ($71.2$\%). This may be because our training set, \emph{im2latexv2}, consists primarily of single-line expressions. The other models suffer a much more dramatic performance drop for multi-line MEs to 14-23\% Edit score. 
Furthermore, by using a straightforward y-cut algorithm, we can strongly improve our model's performance for multi-line MEs from $71.2$\% to $96.2$\% Edit score. As a result, when the y-cut algorithm performs well, we can robustly recognize multi-line MEs even with our model mainly trained on single-line MEs.

\subsubsection{Most Frequent Token Errors}
\label{subsec:problematicTokens}
To better understand the open challenges of our \emph{MathNet} model, we analyzed the Levenshtein operations needed to correct the predictions. Table \ref{tab:tokenIssues} shows the $10$ most frequent tokens that needed be corrected. It is not surprising that the curly brackets are the primary culprit of errors because they are the most frequent tokens in the GT. Also, the sub- and superscript tokens (\verb|_| and \verb|^|) are still tricky for our model, even after our normalization step.\\
The replace operations reveal that the model is mainly confused by visually very similar symbols. However, their occurrences are small compared to the number of errors with curly brackets.

\begin{table*}
    \caption{Analysis of the Levenshtein operations required to correct the \emph{MathNet} predictions on the \emph{im2latexv2} test set. The table shows the $10$ most frequent tokens required to be inserted or deleted and the $10$ most frequent pairs of tokens that must be replaced by the other.}
    \label{tab:tokenIssues}
    \centering
    \begin{tabular}{c|c|c|c|c|c|c}
    \multicolumn{2}{c|}{Insert} & \multicolumn{2}{c|}{Delete} &\multicolumn{3}{c}{Replace}\\
         Token & Count & Token & Count & Token 1 & Token 2 & Count \\
         \hline
         \verb|}| & 2078 & \verb|}| & 666 & \verb|*| & \verb|\ast| & 196\\
         \verb|{| & 2000 & \verb|{| & 589 & \verb|^| & \verb|\star| & 95\\
         \verb|^| & 678 & \verb|_| & 122 & \verb|a| & \verb|\alpha| & 79\\
         \verb|_| & 678 & \verb|*| & 96 & \verb|\nu| & \verb|v| & 74\\
         \verb|2| & 345 & \verb|^| & 63 & \verb|\phi| & \verb|\varphi| & 71\\
         \verb|)| & 344 & \verb|-| & 49 & \verb|\rangle| & \verb|>| & 42\\
         \verb|-| & 271 & \verb|1| & 47 & \verb|\dots| & \verb|\cdots| & 40\\
         \verb|(| & 263 & \verb|2| & 46 & \verb|\epsilon| & \verb|\varepsilon| & 36\\
         \verb|1| & 242 & \verb|&| & 43 & \verb|\langle| & \verb|<| & 34\\
         \verb|,| & 196 & \verb|c| & 42 & \verb|\psi| & \verb|\Psi| & 29\\
    \end{tabular}
\end{table*}

\subsection{Model Related Achievements and Challenges}
\label{subsec:modelRelatedDiscussion}
\begin{figure}
    \centering
    \includegraphics[width=\linewidth]{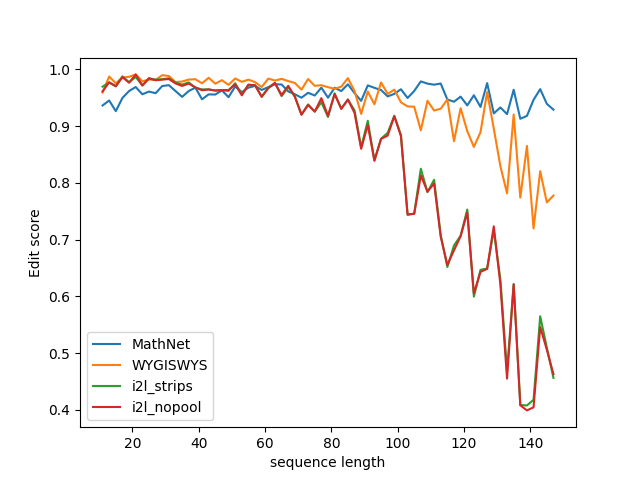}
    \caption{The plot shows the average edit score per sequence length for the different models and the \emph{im2latex-100k} dataset. The x-axis shows the number of tokens in the ME with a bin width of 3. The y-axis shows the average Edit score of each bin. A perfect prediction has a Edit score of 1.}
    \label{fig:editScorePerLength1}
\end{figure}

\begin{figure}
    \centering
    \includegraphics[width=\linewidth]{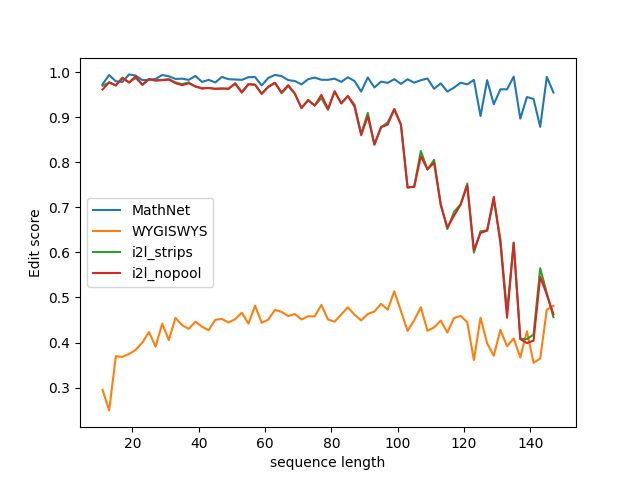}
    \caption{The plot shows the average edit score per sequence length for the different models and the \emph{im2latexv2} dataset. The x-axis shows the number of tokens in the ME with a bin width of 3. The y-axis shows the average Edit score of each bin. A perfect prediction has a Edit score of 1.}
    \label{fig:editScorePerLength2}
\end{figure}

As discussed in Section \ref{sec:RelatedWork}, many MER models employ LSTMs with specialized mechanisms to improve long-distance learning. We addressed this issue using a transformer architecture. Our analysis, depicted in Figures \ref{fig:editScorePerLength1} and \ref{fig:editScorePerLength2}, shows that the Edit score of \emph{MathNet} does not decrease with the sequence length of the MEs, indicating that transformers are effective in learning long-distance relationships in MEs.

\section{Conclusion}
\label{sec:conclusion}
We introduced the novel printed MER model \emph{MathNet}, incorporating a CvT encoder and transformer decoder. \emph{MathNet} achieves
outstanding results for \emph{im2latex-100k} (Edit score: $94.7$\%), \emph{im2latexv2} (Edit score: $97.2$\%), \emph{realFormula} (Edit score: $88.3$\%), and \emph{InftyMDB-1} (Edit score: $89.2$\%), reducing the Edit error rate to the prior state of the art for these datasets by $53.5$\% (from $11.4$\% to $5.3$\%), $88.3$\% (from $24$\% to $2.8$\%), $66.4$\% (from $34.8$\% to $11.7$\%), and $70.4$\% (from $36.5$\% to $10.8$\%), respectively. These results were achieved with our transformer-based model architecture and on an inherently data-centric approach normalizing and augmenting the training data. We found that detrimental variations in the LaTeX GT of \emph{im2latex-100k} exist. To reduce this undesired variations, we proposed a LaTeX normalization method. Our LaTeX normalization process enables the model to focus on the canonical form of an ME instead of learning non-relevant variations. We demonstrated that our LaTeX normalization process is mainly responsible for the model's superior performance. Moreover, we introduced an augmented dataset, \emph{im2latexv2}, an enhanced and normalized version of \emph{im2latex-100k} with multiple fonts, and \emph{realFormula} which contains annotated real ME images from arXiv papers. We also showed that a simple y-cut algorithm can expand single-line MER to multi-line MER.\\
Despite promising effectiveness, the Edit scores of all models investigated were significantly lower on \emph{realFormula} and \emph{InftyMDB-1} compared to \emph{im2latex-100k} and \emph{im2latexv2}, which indicates a difference between synthetic (\emph{im2latex-100k} and \emph{im2latexv2}) and real-world datasets (\emph{realFormula} and \emph{InftyMDB-1}). The removal of mathematical fonts styles in \emph{im2latexv2}, such as bold and italics, limits the correct recognition of MEs that use these mathematical fonts styles in \emph{realFormula}. An extended version of \emph{im2latexv2} with mathematical fonts could solve this problem. Additionally, the correct cutting of ME lines heavily supports multi-line ME recognition, making stable line detectors a precondition.\\
After testing the handwritten benchmark dataset \emph{CROHME} with our model \emph{MathNet} and our LaTeX normalization, we could not find evidence that our LaTeX normalization process helps to improve the recognition performance. We think this is because the characteristics of \emph{CROHME} and \emph{im2latex-100k} are vastly different. The MEs in \emph{CROHME} are on average only one-third as long as in \emph{im2latex-100k}, and the vocabulary is significantly smaller, consisting of only $142$ tokens compared to $500$ tokens in printed MER. As a result, our LaTeX normalization only reduces the original $142$ tokens to $121$ (canonical) tokens, which is much less than with \emph{im2latex-100k}. Furthermore, the MEs in \emph{CROHME} are simpler and do not contain arrays, mathematical fonts, and other complex elements. This leads to the conclusion that the detrimental variation in \emph{CROHME} is much lower than in \emph{im2latex-100k}. However, we believe that for more complex handwritten MEs, our LaTeX normalization process could be as beneficial as it is for printed MER.\\
Generative pretrained transformers with multimodal input have shown significant progress in image recognition. However, testing a few ME images with GPT-4 from OpenAI indicates that the results, although impressive, have not yet reached the state of the art in MER. Nevertheless, combining generative AI with MER could be a promising approach worth exploring.\\
For our upcoming research steps, we plan to combine \emph{FormulaNet} \cite{schmitt-koopmann_formulanet_2022} and \emph{MathNet} to develop a semi-automatic captioning system for MEs in PDFs. With this system, we expect to significantly improve the accessibility of PDFs specifically for MEs and also enable easy searching and extracting of MEs from PDFs.

\bibliography{references}

\begin{thebibliography}{10}

\bibitem{schmitt-koopmann_accessible_2022}
Felix~M. Schmitt-Koopmann, Elaine~M. Huang, and Alireza Darvishy.
\newblock Accessible {PDFs}: {Applying} {Artificial} {Intelligence} for {Automated} {Remediation} of {STEM} {PDFs}.
\newblock In {\em Proceedings of the 24th {International} {ACM} {SIGACCESS} {Conference} on {Computers} and {Accessibility}}, {ASSETS} '22, pages 1--6, New York, NY, USA, 2022. Association for Computing Machinery.
\newblock doi: 10.1145/3517428.3550407.

\bibitem{belaid_syntactic_1984}
Abdelwaheb Belaid and Jean-Paul Haton.
\newblock A {Syntactic} {Approach} for {Handwritten} {Mathematical} {Formula} {Recognition}.
\newblock {\em IEEE Transactions on Pattern Analysis and Machine Intelligence}, PAMI-6(1):105--111, January 1984.
\newblock doi: 10.1109/TPAMI.1984.4767483.

\bibitem{chan_mathematical_2000}
Kam-Fai Chan and Dit-Yan Yeung.
\newblock Mathematical expression recognition: a survey.
\newblock {\em International Journal on Document Analysis and Recognition}, 3(1):3--15, August 2000.

\bibitem{aggarwal_survey_2022}
Ridhi Aggarwal, Shilpa Pandey, Anil~Kumar Tiwari, and Gaurav Harit.
\newblock Survey of {Mathematical} {Expression} {Recognition} for {Printed} and {Handwritten} {Documents}.
\newblock {\em IETE Technical Review}, 39(6):1245--1253, November 2022.
\newblock doi: 10.1080/02564602.2021.2008277.

\bibitem{kanervisto_im2latex-100k_2016}
Anssi Kanervisto.
\newblock im2latex-100k, June 2016.
\newblock doi: 10.5281/zenodo.56198.

\bibitem{zhang_understanding_2021}
Chiyuan Zhang, Samy Bengio, Moritz Hardt, Benjamin Recht, and Oriol Vinyals.
\newblock Understanding deep learning (still) requires rethinking generalization.
\newblock {\em Communications of the ACM}, 64(3):107--115, February 2021.
\newblock doi: 10.1145/3446776.

\bibitem{arpit_closer_2017}
Devansh Arpit, Stanislaw Jastrzebski, Nicolas Ballas, David Krueger, Emmanuel Bengio, Maxinder~S. Kanwal, Tegan Maharaj, Asja Fischer, Aaron Courville, Yoshua Bengio, and Simon Lacoste-Julien.
\newblock A {Closer} {Look} at {Memorization} in {Deep} {Networks}.
\newblock In {\em Proceedings of the 34th {International} {Conference} on {Machine} {Learning}}, pages 233--242. PMLR, July 2017.

\bibitem{deeplearningai_chat_2021}
{DeepLearningAI}.
\newblock A {Chat} with {Andrew} on {MLOps}: {From} {Model}-centric to {Data}-centric {AI}, March 2021.
\newblock [Online]. Available: https://www.youtube.com/watch?v=06-AZXmwHjo.

\bibitem{luley_concept_2023}
Paul-Philipp Luley, Jan~M. Deriu, Peng Yan, Gerrit~A. Schatte, and Thilo Stadelmann.
\newblock From {Concept} to {Implementation}: {The} {Data}-{Centric} {Development} {Process} for {AI} in {Industry}.
\newblock In {\em 2023 10th {IEEE} {Swiss} {Conference} on {Data} {Science} ({SDS})}, pages 73--76, June 2023.
\newblock doi: 10.1109/SDS57534.2023.00017.

\bibitem{stadelmann_data_2022}
Thilo Stadelmann, Tino Klamt, and Philipp~H. Merkt.
\newblock Data {Centrism} and the {Core} of {Data} {Science} as a {Scientific} {Discipline}.
\newblock {\em Archives of Data Science, Series A}, 8(2):16, 2022.
\newblock doi: 10.5445/IR/1000143637.

\bibitem{anderson_syntax-directed_1967}
Robert~H. Anderson.
\newblock Syntax-directed recognition of hand-printed two-dimensional mathematics.
\newblock In {\em Symposium on {Interactive} {Systems} for {Experimental} {Applied} {Mathematics}: {Proceedings} of the {Association} for {Computing} {Machinery} {Inc}. {Symposium}}, pages 436--459, New York, NY, USA, August 1967. Association for Computing Machinery.
\newblock doi: 10.1145/2402536.2402585.

\bibitem{xie_icdar_2023}
Yejing Xie, Harold Mouchère, Foteini Simistira~Liwicki, Sumit Rakesh, Rajkumar Saini, Masaki Nakagawa, Cuong~Tuan Nguyen, and Thanh-Nghia Truong.
\newblock {ICDAR} 2023 {CROHME}: {Competition} on {Recognition} of {Handwritten} {Mathematical} {Expressions}.
\newblock In Gernot~A. Fink, Rajiv Jain, Koichi Kise, and Richard Zanibbi, editors, {\em Document {Analysis} and {Recognition} - {ICDAR} 2023}, Lecture {Notes} in {Computer} {Science}, pages 553--565, Cham, 2023. Springer Nature Switzerland.
\newblock doi: 10.1007/978-3-031-41679-8\_33.

\bibitem{suzuki_infty_2003}
Masakazu Suzuki, Fumikazu Tamari, Ryoji Fukuda, Seiichi Uchida, and Toshihiro Kanahori.
\newblock {INFTY}: an integrated {OCR} system for mathematical documents.
\newblock In {\em Proceedings of the 2003 {ACM} symposium on {Document} engineering}, {DocEng} '03, pages 95--104, New York, NY, USA, November 2003. Association for Computing Machinery.
\newblock doi: 10.1145/958220.958239.

\bibitem{malon_mathematical_2008}
Christopher Malon, Seiichi Uchida, and Masakazu Suzuki.
\newblock Mathematical symbol recognition with support vector machines.
\newblock {\em Pattern Recognition Letters}, 29(9):1326--1332, July 2008.
\newblock doi: 10.1016/j.patrec.2008.02.005.

\bibitem{deng_image--markup_2017}
Yuntian Deng, Anssi Kanervisto, Jeffrey Ling, and Alexander~M. Rush.
\newblock Image-to-{Markup} {Generation} with {Coarse}-to-{Fine} {Attention}.
\newblock In {\em Proceedings of the 34th {International} {Conference} on {Machine} {Learning} - {Volume} 70}, {ICML}'17, pages 980--989, Sydney, NSW, Australia, August 2017. JMLR.org.

\bibitem{hochreiter_long_1997}
Sepp Hochreiter and Jürgen Schmidhuber.
\newblock Long {Short}-{Term} {Memory}.
\newblock {\em Neural Computation}, 9(8):1735--1780, November 1997.
\newblock doi: 10.1162/neco.1997.9.8.1735.

\bibitem{cho_learning_2014}
Kyunghyun Cho, Bart van Merriënboer, Caglar Gulcehre, Dzmitry Bahdanau, Fethi Bougares, Holger Schwenk, and Yoshua Bengio.
\newblock Learning {Phrase} {Representations} using {RNN} {Encoder}–{Decoder} for {Statistical} {Machine} {Translation}.
\newblock In {\em Proceedings of the 2014 {Conference} on {Empirical} {Methods} in {Natural} {Language} {Processing} ({EMNLP})}, pages 1724--1734, Doha, Qatar, October 2014. Association for Computational Linguistics.
\newblock doi: 10.3115/v1/D14-1179.

\bibitem{vaswani_attention_2017}
Ashish Vaswani, Noam Shazeer, Niki Parmar, Jakob Uszkoreit, Llion Jones, Aidan~N Gomez, Lukasz Kaiser, and Illia Polosukhin.
\newblock Attention is {All} you {Need}.
\newblock In {\em Advances in {Neural} {Information} {Processing} {Systems}}, volume~30, pages 1--11. Curran Associates, Inc., 2017.

\bibitem{cho_properties_2014}
Kyunghyun Cho, Bart van Merriënboer, Dzmitry Bahdanau, and Yoshua Bengio.
\newblock On the {Properties} of {Neural} {Machine} {Translation}: {Encoder}–{Decoder} {Approaches}.
\newblock In {\em Proceedings of {SSST}-8, {Eighth} {Workshop} on {Syntax}, {Semantics} and {Structure} in {Statistical} {Translation}}, pages 103--111, Doha, Qatar, October 2014. Association for Computational Linguistics.
\newblock doi: 10.3115/v1/W14-4012.

\bibitem{bian_handwritten_2022}
Xiaohang Bian, Bo~Qin, Xiaozhe Xin, Jianwu Li, Xuefeng Su, and Yanfeng Wang.
\newblock Handwritten {Mathematical} {Expression} {Recognition} via {Attention} {Aggregation} {Based} {Bi}-directional {Mutual} {Learning}.
\newblock {\em Proceedings of the AAAI Conference on Artificial Intelligence}, 36(1):113--121, June 2022.
\newblock doi: 10.1609/aaai.v36i1.19885.

\bibitem{li_when_2022}
Bohan Li, Ye~Yuan, Dingkang Liang, Xiao Liu, Zhilong Ji, Jinfeng Bai, Wenyu Liu, and Xiang Bai.
\newblock When {Counting} {Meets} {HMER}: {Counting}-{Aware} {Network} for {Handwritten} {Mathematical} {Expression} {Recognition}.
\newblock In {\em Computer {Vision} – {ECCV} 2022: 17th {European} {Conference}, {Tel} {Aviv}, {Israel}, {October} 23–27, 2022, {Proceedings}, {Part} {XXVIII}}, pages 197--214, Berlin, Heidelberg, 2022. Springer-Verlag.
\newblock doi: 10.1007/978-3-031-19815-1{\textbackslash}\_12.

\bibitem{yan_convmath_2021}
Zuoyu Yan, Xiaode Zhang, Liangcai Gao, Ke~Yuan, and Zhi Tang.
\newblock {ConvMath}: {A} {Convolutional} {Sequence} {Network} for {Mathematical} {Expression} {Recognition}.
\newblock In {\em 2020 25th {International} {Conference} on {Pattern} {Recognition} ({ICPR})}, pages 4566--4572, Milan, Italy, January 2021. IEEE.
\newblock doi: 10.1109/ICPR48806.2021.9412913.

\bibitem{mirkazemy_mathematical_2023}
Abolfazl Mirkazemy, Peyman Adibi, Seyed Mohhamad~Saied Ehsani, Alireza Darvishy, and Hans-Peter Hutter.
\newblock Mathematical expression recognition using a new deep neural model.
\newblock {\em Neural Networks}, 167:865--874, October 2023.
\newblock doi: 10.1016/j.neunet.2023.08.045.

\bibitem{wang_image_2019}
Jian Wang, Yunchuan Sun, and Shenling Wang.
\newblock Image {To} {Latex} with {DenseNet} {Encoder} and {Joint} {Attention}.
\newblock {\em Procedia Computer Science}, 147:374--380, 2019.
\newblock doi: 10.1016/j.procs.2019.01.246.

\bibitem{li_improving_2020}
Zhe Li, Lianwen Jin, Songxuan Lai, and Yecheng Zhu.
\newblock Improving {Attention}-{Based} {Handwritten} {Mathematical} {Expression} {Recognition} with {Scale} {Augmentation} and {Drop} {Attention}.
\newblock In {\em 2020 17th {International} {Conference} on {Frontiers} in {Handwriting} {Recognition} ({ICFHR})}, pages 175--180. IEEE Computer Society, September 2020.
\newblock doi: 10.1109/ICFHR2020.2020.00041.

\bibitem{peng_image_2021}
Shuai Peng, Liangcai Gao, Ke~Yuan, and Zhi Tang.
\newblock Image to {LaTeX} with {Graph} {Neural} {Network} for {Mathematical} {Formula} {Recognition}.
\newblock In {\em Document {Analysis} and {Recognition} – {ICDAR} 2021: 16th {International} {Conference}, {Lausanne}, {Switzerland}, {September} 5–10, 2021, {Proceedings}, {Part} {II}}, pages 648--663, Berlin, Heidelberg, September 2021. Springer-Verlag.
\newblock doi: 10.1007/978-3-030-86331-9\_42.

\bibitem{mahdavi_icdar_2019}
Mahshad Mahdavi, Richard Zanibbi, Harold Mouchere, Christian Viard-Gaudin, and Utpal Garain.
\newblock {ICDAR} 2019 {CROHME} + {TFD}: {Competition} on {Recognition} of {Handwritten} {Mathematical} {Expressions} and {Typeset} {Formula} {Detection}.
\newblock In {\em 2019 {International} {Conference} on {Document} {Analysis} and {Recognition} ({ICDAR})}, pages 1533--1538, September 2019.
\newblock doi: 10.1109/ICDAR.2019.00247.

\bibitem{singh_teaching_2018}
Sumeet~S. Singh.
\newblock Teaching {Machines} to {Code}: {Neural} {Markup} {Generation} with {Visual} {Attention}, June 2018.
\newblock doi: 10.48550/arXiv.1802.05415.

\bibitem{blecher_nougat_2023}
Lukas Blecher, Guillem Cucurull, Thomas Scialom, and Robert Stojnic.
\newblock Nougat: {Neural} {Optical} {Understanding} for {Academic} {Documents}, August 2023.
\newblock doi: 10.48550/arXiv.2308.13418.

\bibitem{simmler_survey_2021}
Niclas Simmler, Pascal Sager, Philipp Andermatt, Ricardo Chavarriaga, Frank-Peter Schilling, Matthias Rosenthal, and Thilo Stadelmann.
\newblock A {Survey} of {Un}-, {Weakly}-, and {Semi}-{Supervised} {Learning} {Methods} for {Noisy}, {Missing} and {Partial} {Labels} in {Industrial} {Vision} {Applications}.
\newblock In {\em 2021 8th {Swiss} {Conference} on {Data} {Science} ({SDS})}, pages 26--31, June 2021.

\bibitem{schmitt-koopmann_formulanet_2022}
Felix~M. Schmitt-Koopmann, Elaine~M. Huang, Hans-Peter Hutter, Thilo Stadelmann, and Alireza Darvishy.
\newblock {FormulaNet}: {A} {Benchmark} {Dataset} for {Mathematical} {Formula} {Detection}.
\newblock {\em IEEE Access}, 10:91588--91596, 2022.
\newblock doi: 10.1109/ACCESS.2022.3202639.

\bibitem{dosovitskiy_image_2021}
Alexey Dosovitskiy, Lucas Beyer, Alexander Kolesnikov, Dirk Weissenborn, Xiaohua Zhai, Thomas Unterthiner, Mostafa Dehghani, Matthias Minderer, Georg Heigold, Sylvain Gelly, Jakob Uszkoreit, and Neil Houlsby.
\newblock An {Image} is {Worth} 16x16 {Words}: {Transformers} for {Image} {Recognition} at {Scale}.
\newblock In {\em 9th {International} {Conference} on {Learning} {Representations}, {ICLR} 2021, {Virtual} {Event}, {Austria}, {May} 3-7, 2021}, pages 1--21. OpenReview.net, 2021.

\bibitem{wu_cvt_2021}
Haiping Wu, Bin Xiao, Noel Codella, Mengchen Liu, Xiyang Dai, Lu~Yuan, and Lei Zhang.
\newblock {CvT}: {Introducing} {Convolutions} to {Vision} {Transformers}.
\newblock {\em 2021 IEEE/CVF International Conference on Computer Vision (ICCV)}, pages 22--31, October 2021.
\newblock doi: 10.1109/ICCV48922.2021.00009.

\bibitem{kingma_adam_2014}
Diederik~P. Kingma and Jimmy Ba.
\newblock Adam: {A} {Method} for {Stochastic} {Optimization}.
\newblock In {\em {CoRR}}, pages 1--13. Ithaca, NY: arXiv.org, December 2014.
\newblock doi: https://hdl.handle.net/11245/1.505367.

\bibitem{fujiyoshi_inftymdb-1_2009}
Akio Fujiyoshi, Masakazu Suzuki, and Seiichi Uchida.
\newblock {InftyMDB}-1, 2009.
\newblock [Online]. Available: https://www.inftyproject.org/download/inftydb/InftyMDB-1.zip.

\end{thebibliography}

\end{document}